# The method of artificial systems

*Christopher A. Tucker, Ph.D., Cartheur Technology Research, Delft, The Netherlands*

"I hope that once we have succeeded in making computer programs reason about the world, we will be able to reformulate epistemology as a branch of applied mathematics no more mysterious or controversial than physics."

–John McCarthy

Abstract

This document is written with the intention to describe in detail a method and means by which a computer program can reason about the world and in so doing, increase its analogue to a living system. As the literature is rife and it is apparent we, as scientists and engineers, have not found the solution, this document will attempt the solution by grounding its intellectual arguments within tenets of human cognition in Western philosophy. The result will be a characteristic description of a method to describe an artificial system analogous to that performed for a human. The approach was the substance of my Master's thesis, explored more deeply during the course of my postdoc research. It focuses primarily on context awareness and choice set within a boundary of available epistemology, which serves to describe it. Expanded upon, such a description strives to discover agreement with Kant's critique of reason to understand how it could be applied to define the architecture of its design. The intention has never been to mimic human or biological systems, rather, to understand the profoundly fundamental rules, when leveraged correctly, results in an artificial consciousness as noumenon while in keeping with the perception of it as phenomenon. The natural progression, then, would be to discover what would be the substance of its implementation. As it is also interesting to determine the sense of will—as a clarification to the transcendental aesthetic, the concept of body as feedback mechanism will be discussed as to what influence upon the design there may be. Ultimately, this document will demonstrate using such techniques will allow an empirical ethically-traceable framework where behaviour eccentricities can be known in advance by classification of pathway logic and attenuated on a scale such that a formal definition as to its type can be revealed. The revelation is then, in itself, a rational system.

## 1 Introduction

This document proposes a solution to conundrums within artificial intelligence which states that machines which possess an aspect exhibited by living analogues cannot be designed by making assumptions as to their psychological consistence; rather, they must be designs derivative from a template of the mind and consciousness in a general sense—ideally from a parametric set of rules governing pathways through states, mitigating experience as a temporal flow of information from the environment. Although numerous forms can claim to exist, a proper architecture needs to be authored to establish baselines for generational development of both the form and function of the artificial intelligence. Ultimately, the aim would be to provide the program the ability to learn from experience as effectively as humans do whereby creating an understanding about context in the program. This, then, allows the creation and alteration of scenarios, or behavior templates, that are anticipated by functions in the code.

In order to set a traceable foundation for this, the template will be derived from Kant [1] and Schopenhauer [2] treatises on reason and metaphysics. The former for the definition of mind, consciousness, experience, and understanding in cognition and the latter regarding corrections of perception in the transcendental analytic where the notion where sense impressions and not objects are given by sensibility. It is hypothesized that the creation and implementation of such a template in software (and hardware where appropriate) yields a strong coupling of an ethical foundation, allowing consequential adaptation mechanism for an artificial entity—be it in the form of a computer program, robot, or distributed system—to manifest within the program's runtime. The problem is presented by discussing different points of view and proposing experiments to demonstrate the complexities of general problems of machine consciousness while indicating what a suitable answer might be.

Aims

The four primary aims of this document are:

1. To describe a methodological, epistemological, behavioral, and experimental foundation to understand and construct autonomous synthetic creatures of varying purpose. Directed by the application environment, machine types form an active role of technological development whose anticipated psychological "state of mind" should be classified,
2. to propose a research paradigm to understand artificial entities from a psychological perspective, and perhaps shed some light on what is life in a more general sense, by designing and constructing myriad forms, and,
3. to propose experiments which convey motifs of behavior, to establish a paradigm set of discretionary limits, necessitated by a physical characteristic,
4. of the agenda to quantify artificial intelligence by classifying behaviors identified for control to be included within a desired ethical foundation.

The seemingly dubious intent appearing as its final form within the paradigm of a practical engineering point-of-view will provide unique and useful solutions how to design, construct, control, and regulate the operation of autonomous machines possessing ranges of adapted, organically-developed behaviors. This is a keen individual experience, which comes to bear on the outcomes and must be taken in *a posteriori* analysis by the reader.

Terms

This document will juxtapose loosely defined concepts nested in objectifying terms such as *robot*, *machine*, *autonomous*, *synthetic creature*, *artificial life form*, and the like to describe heretofore abstractly realized phenomenon human consciousness compartmentalizes. To carry a specific and targeted meaning of the abstract concepts that will be introduced and explored herein, terms used widely throughout are:

Feature: the prominent part or characteristic embedded in both the physical and abstract components.

Entity: a dynamic object that possesses behavior forming a commonly identifiable form.

Robot: interchangeably used with *machine*, *autonomous entity*, and *synthetic creature* meaning a composite technical form who may or may not bear resemblance to living forms, divorced from its original context or exact translation.

Controller: an entity that monitors categorized machine states based on external criteria.

Orchestration: cooperative coordination between disparate physical or abstract elements.

Autonomy: A system that exhibits behavior derived from experience or evolution in its programming.

The terms *life* and *living* are employed loosely throughout this document. They convey to describe *any natural or synthesized system that interacts directly with a human observer ascribing interest in it as an object*.

Literature

This document derives its intellectual arguments from the contributions of Ashby, Craik, Kant, Lighthill, McCarthy, Schopenhauer, and Walter.

## 2  A.I. as a general sensibility

A brief history of artificial intelligence stretches back to the first doctrines of logic by Plato and personage by Aristotle and the development of mathematics by luminaries such as Archimedes and the unnamed engineers of Rhodes whose contributions were unknown until the 20$^{th}$ Century. Researchers McCarthy, Von Neumann, and others compartmentalized this notion, which facilitated a rife of experimentation well into the 21$^{st}$ Century. In its current form, a critical review of A.I. will yield reluctance by those considered expert in the field [3]. An analysis *quid pro quo* indicates intractable problems such as knowledge (epistemological) representation and choice-motivated (transcendental) decision (aesthetic) making (analytic) are beyond the scope of current understanding to solve the A.I. problem.

Autonomous machines, which are made to be independent, work collectively yet imbued with the need of human interaction to *survive*, infer a representation, which could be identified as purpose. A direct consequence of this, is the consideration that the machine possesses an *inner* world of metaphor constructed similarly as in humans but radically different given the geometry of its experience. While there are varying arguments for human-attributed A.I., such as conversational agents, and non-human attributed A.I., such as a robotic interaction using tones or inferred physical (visual) cues, regardless each can be brought into a single sensibility sufficient to answer engineering questions.

Notwithstanding a general description, living analogues serve as inspirational sources where the notion of choice in the machine is a quantifier for pure representation.

Some questions posited on living systems

How can the events in space and time, which take place within the spatial boundary of a living organism, be accounted for in terms of physics and chemistry? It is reasonable then, to look toward biological analogues and the work in the biological and ethological fields wherein to derive hypotheses with a high degree of reliability and purpose. How is this information compiled? In other words, what is life *quid pro* *quo* whose meaning is gleaned from experience functioning in an environment? Does the phenomenon we observe and compartmentalize to be identified as a living system result from the properties of a noumenon called the brain or does it exist in a very general way, in different kinds of brains that are understood through continued experience over time?

Why should an organ like a brain, with the sensorial system attached to it, of necessity consist of an enormous number of neurons, in order that its physically changing state should be in close and intimate correspondence with a highly developed thought? On what grounds is the latter task of the said organ incompatible with being, as a whole or in some of its peripheral parts, which interact directly with the environment, a mechanism sufficiently refined and sensitive to respond to and register the impact of a single neuron? The reason for this is, that what is called thought (1) is itself an orderly thing, and (2) can only be applied to material, i.e. to perceptions or experiences, which have a certain degree of orderliness. This has two consequences. First, a physical organization, to be in close correspondence with thought—as of brain by thought—must be a very well ordered organization, and that means that the events that happen within it must obey strict physical laws, at least to a very high degree of accuracy. Secondly, the physical impressions made upon that physically well-organized system by other bodies from outside, obviously correspond to the perception and experience of the corresponding thought, forming its material. Therefore, the physical interactions between our system and others must as a matter of trait, possess a certain degree of physical orderliness, that is to say, they too must obey strict physical laws to a certain degree of accuracy.

Therefore, a creature with a set of rigorous functionality, e.g., suited to its environment to a reasonable approximation of regularity or possessing a homeostatic point [4]. When the dynamic system can vary continuously, small disturbances are, in practice, usually acting on it incessantly. Electronic systems are disturbed by thermal agitation, mechanical systems by vibration, and biological systems by a host of minor disturbances. For this reason, the only states of equilibrium that can persist are those that are stable in the magnitude difference of its invariance to stability. States of unstable equilibrium are of small practical importance in the continuous system though they may be of importance in the system that can change only by a discrete jump. The concept of unstable equilibrium is, however, of some theoretical importance.

Which shows how physical laws are expressed internally and externally to the entity considered to be living, with an acceptable range of what causes can be known and compensated for (stable) and what causes remain unknown (unstable). This implies the physical phenomena of order, disorder, and entropy all play a role in the living system whether organic or technic. In a conceptually analytic manner, life seems to be orderly and lawful behaviour of matter, not based exclusively on its tendency to go over from order to disorder, but based partly on existing order that is kept up. The living organism is a macroscopic system, which in part of its behaviour approaches to a purely mechanical conductivity (as contrasted with thermodynamic in organic representation) to which all systems tend.

It is by avoiding the rapid decay into the inert state of equilibrium that an organism appears so enigmatic; so much so, that from the earliest times of human thought some special non-physical or supernatural forcive *entelechy* is the noumenon responsible for this class of phenomena. How does the living organism avoid decay, or more mathematically, reverse entropy? Where for the purposes of this document metabolism for an artificial system is the exchange of information and growth of its program complex in the most rudimentary terms of the concept, inclusive of the wear on physical components such as circuits, memory media, lights and motors—following Schopenhauer—illustrating the artificial system is privy to the forces of entropy the same as a living one would. As the entity co-exists with its environment, order is maintained by the input flow of information–negative entropy–creating a balance or state of theoretical equilibrium that could be calculated for a given artificial entity. The reality of this realization is that even if an artificial system is thought of as a purely mechanical thing, devoid of even the intrinsic bits of philosophically imbued life, the probability of its existing outside of its definition is low.

Epistemology

A most fundamental viewpoint of this work is epistemology; without a framing the discussion, that is, an agreed framework of how new concepts will be discussed and how they will be manifest empirically, we will never arrive at quantitative proofs of artificial life. Foundational controversies in artificial life and artificial intelligence arise from lack of decidable criteria for defining the epistemic cuts that separate knowledge of reality from supposed reality itself, e.g., description from construction, simulation from realization, and mind from brain. When a problem persists, unresolved...in spite of enormous increases in our knowledge, it is a good bet that the problem entails the nature of knowledge itself [5], understood as an oscillation between the limits of two epistemic cuts:

- *Life-as-it-could-be* compared with
- *Life-as-we-know-it*.

These cuts illustrate the problem space in terms of control in the evolution at runtime of the formation of a synthetic life form defined by an *a posteriori* analytic of empirical data. The illustration includes:

1. Intellectual activity takes place in a world that has a certain physical and intellectual structure: Physical objects exist, move about, are created and destroyed. Actions that may be performed have effects that are partially known. Entities with goals have available to them certain information about the world. Some of this information may be built in, and some arises from observation, from communication, from reasoning, and by more or less complex processes of retrieval from information bases. Much of this structure is common to the intellectual position of animals, people, and machine, which we may design, e.g. the effects of physical actions on material objects and the information that may be obtained about these objects by vision. The general structure of the intellectual world is far from understood, and it is often quite difficult to decide how to represent effectively the information available about a quite limited domain of action even when we are quite willing to treat a particular problem in an ad hoc way.
2. The process of problem solving depends on the class of problems being solved more than on the solver. Thus, playing chess seems to require look-ahead whether the apparatus is made of neurons or transistors. Isolation of the information relevant to a problem from the totality of previous experience is required whether the solver is man or machine, and so is the ability to divide a problem into weakly connected sub-problems that can be thought about separately before the results are combined.

3. Experiment is useful in determining what representations of information and what problem solving processes are needed to solve a given class of problems.
4. The experimental domain should be chosen to test the adequacy of representations of information and of problem solving mechanisms as the Drosophila of artificial intelligence—where A.I. as performance criteria is how well it plays strategic, rule-based games such as Mahjong, chess, Sudoku, and Go. Such criteria require the ability to identify, represent, and recognize patterns of position and play that correspond to optimal ideals within the problem domain of abstract to actual positions of rule-based play [6].

This document will now attempt to derive an epistemology, a way of speaking about the subject and a manner of defining experiments–coupled with the attempt to ascertain a reasonable model of artificial intelligence as exhibited in form analogous with a living system.

A model for consideration

When establishing the consideration for a singular model of A.I., ascribed by this document to contain even a simple set of parameters analogous to a living system, a few targeted examples exist. The work of W. Grey Walter is particularly useful as a means to understand machine behavior based on physical models of the brain [7]. Through the lens of cybernetics, living systems are organized and how they are related to each other. What is gleaned by this type of categorization of artificial life forms with their biological analogues is the principled interest in building physical working models to test hypotheses, which yield an a posteriori understanding of behavior in terms of brain architecture.

His strength was in the position that the machines, Elmer and Elsie, possessed behaviors unaccounted for in his theory, i.e. emergence. He also expressed the concept of free will observed in his experiments in the psychological concept of the *free goal-seeking mechanism* first mentioned by Craik [8]. Holland [9] provides an analysis into Walter's machines.

> "I knew that in 1949 Grey Walter had built a robot to demonstrate his ideas about how the brain worked. He did not think humans were intelligent just because they had ten billion brain cells, but rather because their brain cells were connected up in many different ways. So he built his first 'model animal'–Elmer the tortoise–using only two electronic brain cells, connected together in several different ways. Not much of a brain, but Grey Walter had designed it very cleverly. Elmer would explore a room, looking for lights, moving towards them, circling them, and then wandering off in search of more. If he found a mirror, he would do a dance in front of it; if he came across his sister, Elsie, he would dance with her. If he came across an obstacle he would try and push it out of the way; if this didn't work, he would go round it. And when his battery began to run down, he would return to his hutch, and plug himself in to his power socket, setting off again in search of lights when his battery was fully charged. Grey Walter had proved his point–two richly connected brain cells were enough."

What is striking here is the intention of Walter to reproduce, by modeling, living systems and that emergent behavior was observed by the utility of rather simple parts in what became known as biologically-inspired robotics. The tortoises were designed to test a hypothesis about how combinations of relatively few elements might give rise to complexity of behavior. The robots were intended to produce behaviour characteristic of animals in terms of behavioral completeness [7]:

> "Not in looks, but in action, the model must resemble an animal. Therefore, it must have these or some measure of these attributes: exploration, curiosity, free-will in the sense of unpredictability, goal-seeking, self-regulation, avoidance of dilemmas, foresight, memory, learning, forgetting, association of ideas, form recognition, and the elements of social accommodation. Such is life."

The tortoises had to exist in a normal everyday environment, rather than in some special environment created to take account of their limitations. He was the first to implement a self-recharging robot. He made the first observations of emergence in robotics, both in the sense of the designer being pleasantly surprised at the unanticipated appearance of some useful side effect of his design, and in the sense that the interaction of two or more behavioral subsystems could produce a distinct and useful additional behaviour. The second sense is clearly demonstrated by several of his remarks in Walter (1960); in fact, they amount to the earliest formulation of the basic idea of behavior-based robotics. As noted above, he was the first to show how a robot's actions on an environment could change it in such a way that the robot's future behavior was changed in a useful way, and he was the first to carry out experiments in learning on a behaving robot. Because he built more than one robot, he was also the first in the field of multiple robotics, showing how the behavioral interactions between two robots of the same type would produce emergent characteristics of interest if not utility. He also made the earliest observations in the field of what is now known as collective robotics [10]:

> "Simple models of behaviour can act as if they could recognize themselves and one another; furthermore, when there are several together they begin to aggregate in pairs and flocks, particularly if they are crowded into a corral. . . . The process of herding is nonlinear. In a free space they are individuals; as the barriers are brought in and the enclosure diminishes, suddenly there is a flock. But if the crowding is increased, suddenly again there is a change to an explosive society of scuffling strangers. And at any time the aggregation may be turned into a congregation by attraction of all individuals to a common goal. Further studies have shown that in certain conditions one machine will tend to be a *leader*. Often this one is the least sensitive of the crowd, sometimes even it is *blind*."

What can be gleaned from selecting as a model for experimentation, Walter's work and Holland's analysis of it and its impact on the current theme? A specific example would be the following: If one were to set out from scratch to build a mechanical creature, such as an owl, it can be defined that it should be designed as:

1. To behave like its natural analogue in a close an approximation as possible, and,
2. allow the machine to develop independent behaviors as a consequence of its own experience by making choices, while simultaneously serving to give insight into how an artificial system would evolve in its "genetic" isolation while cooperating with other entities in an environment within the range of its experiential ability. The dynamics of choice and by quantifying them, a series of study can unfold to get at least the smallest definition of free will and what it means to artificial systems and their adaptation strategies in a tangible environment in the operational paradigm described and used by Craik and Walter.

Up to this point in the document, it has set the foundations in a finite set of influences, demonstrating the connection and impact of those works. The remaining part of the document will discuss the derivation that has been achieved by applying these works in a method and means to design an architecture, which is complete and descriptive.

## 3  The method of artificial systems

This section will begin with the basic premise that nothing of relevance to the topic of this document can be obtained absent of a cohesive method of expression that will guide all subsequent depictions and models derived as such. This document makes a series of arguments designed to illustrate the inherent problem with analysis of artificial systems. The first argument is if a machine is given the ability to make choices and as a consequence of those choices, possesses the feature of the direct experience of entropy, it will foster emergence, manifest behaviors not accounted for in theory or experiment. Choice is the key word and will be used as an epistemological foundation for theoretical and experimental manifestation of what is proposed in this document. It is an expression of the method that the manifestation of the machine in independent articulated form should be autonomous and not privy to interference in learning categories and behaviors from an active outside entity. It consists of three methodological assumptions and the resultant questions that are firstly derived from them.

Method #1: If a machine can be made to mimic biological life, it has access to those kind of experiences attributed to it.

Result #1: Will the empirical ability to make a choice and possess knowledge of it foster richer behavior motifs?

Method #2: If a machine can make a choice, it portrays some level of consciousness.

Result #2: Does the notion of free will and the power to make choices foster a rudimentary sense of self-awareness?

Method #3: If a machine has knowledge of its death, it fosters emergent behavior.

Result #3: Does the placing of a polar opposite to life preclude an organism to conspire against it?

Neither can be absolutely quantified, where possible an arbitrary qualification by juxtaposition with established tenants such as those realized in ethology, fuzzy logic, finite state automata, and self-constructors. It is the aim, then, to derive a suitable algorithmic template by which artificial beings can be realized in technology, a set of laws can be authored, and wherein future generations can be established based on a basic by *foundation*.

Details of the method

The tenets of biologically-inspired robotics are founded on the principle of robots that mimic typified life forms. The foundation is divided into two methodological principles of: 1. attempting to create functional robots based on living systems, and, 2. creating robots to understand biological systems. A deeper examination of these principles yields the first fundamental method.

Method #1: If a machine mimics biological life, it has access to those kinds of experiences attributed to it.

The arguments in this document are engaged in the first principle since the problem motivation is focused on creating functional, autonomous robots who can subsist for themselves in an environment free from human intervention. To this end, the machine must resemble a natural form and function in some sense of the words to make it recognizable when viewed by a participant. In order to necessitate forms and functions, a typography of life forms is required to ascribe relevant behavior necessary to the type of machine desired, an arbitrary classification to set a foundation is shown in Fig.1.

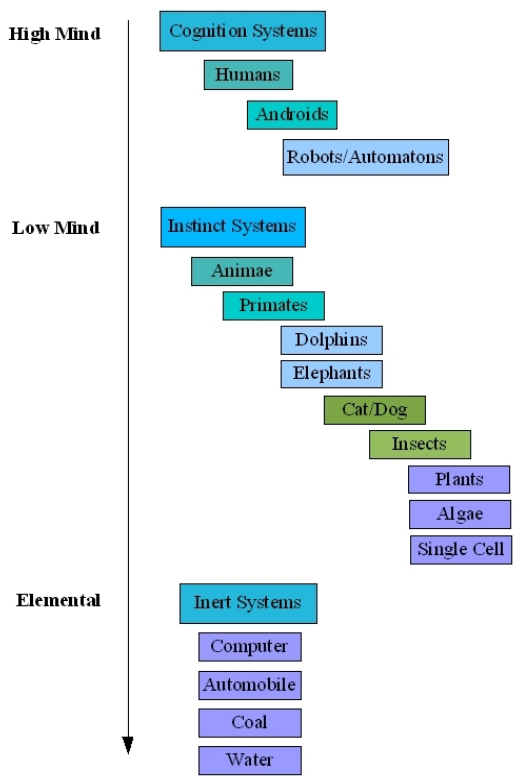

Fig.1. A cognition hierarchy.

The hierarchy of available objects comprising a planetary body, in this instance a closed system. It is organized by decreasing emphasis in ascribed intelligence. The continuum of life is divided into three major categories: Cognition, Instinct, and Inert. Within the categories are sub-categories contrasted by their inherent complexity. For example, in the cognition group, humans are at the top of the hierarchy because of their contrasting complexity to the other members in the group. Androids are lesser as they are designed to mimic humans but are not necessarily privy to the level of complexity manifest in humans dependent upon the level of technological development. Robots and automatons fill the rest of the category, as they are less complex than androids but more complex than the members of the instinct group are. At the top of the hierarchy of instinct systems are animae, these are synthesized animals. Ideally and dependent upon the cleverness of the programmer, they are more complex than the other members of its group are. Animae can be in the form of cats, dogs, or other synthetic animals. Animae is a noninvasive place to start investigating questions of the properties of what constitutes a synthetic life form; they are complex enough to inspire interesting research questions. Currently in artificial intelligence, these pursuits are done in instinct systems, which may or may not yield interesting enough results for purposes of artificial life. However, it still may be of value, high enough in the list to try to understand agents, particulates of intelligence, as a function of automata entities in computing environments. It should be noted that a third system does exist within the frame of this definition, the inert. In a physical sense, the inert plays only a peripheral role to both systems; however, in an abstract sense, the inert plays a crucial role in establishing a closed-loop context. It aids in ascribing and formulating behavior and choice strategies of the other two systems. In such a view, it is a third party. Intelligence is ascribed as a sub-property of each system and arranged in a hierarchical form by the entities that fill each of them. This arrangement is by no means authoritative but serves to illustrate the difference between organic and inorganic systems and where the concept of life may be empirically understood.

For the purposes of the research, this document will assume that cognition is a state of self-actualization. Humans know this phenomenon only individualized experience and can only approximate the experience of others, including other types of living systems. This, however, should not prevent a broader definition since the degree of approximation is limited by the definition itself. Firstly, there is a line of demarcation between a cognitive system and an instinctive system. A cognitive system is the result of a complex evolution of an individual's experience and an instinctive system is the result of the propagation of genetic information between generations of living entities. Secondly, a cognitive system has the ability to extend beyond the quantity of genetic information of an instinctive system and expresses a qualitative difference. Thirdly, this qualitative difference gives rise to two equal yet distinct states of being—the high mind and the low mind. Therefore, cognition is the ability to direct changes on continuous events. Instinct is the inability to affect cognitively the outcome of continuous events. Although there are anomalies that may traverse the definition dependent upon empathetic qualification, it nevertheless holds for most entities.

Method #2: If a machine can make a choice, it portrays some level of consciousness.

The question regarding choice is accepting whether can a machine make a choice and if it did, would it be a meaningful one. On the condition that artificial life is no different that organic life, if an organism is dependent upon its survival and purpose, i.e., obeys the law of entropy, then it must by default make choices regarding the success or failure of its species. This mechanism is the characteristic theory of evolution. If a species could not make the proper choices, adapt to changing conditions, and then it will become extinct. This feature needs to be extended to artificial life forms to see if the quality of artificial life is real or an imagined property. The only way to know whether the forms are alive and prone to the forces of evolution. Architecture should be designed that sets the purpose of the machine and its behavior as an emergent property exhibited in its choices.

How is behavior generated from architecture, how can the intellectual link be made? Introducing a programmatic construct based around the combinatorial philosophy of Kant and Schopenhauer facilitates an intellectual pathway to empirically test the concreteness of a living, finite system comprised of a series of experiences thorough choices creating the domain of understanding of a given problem space. Employing this construct to synthetic machines for if, they are to be considered truly alive, and then they must obey the basic condition that they can die and that they have knowledge of it.

In the specific instance of the noumenon of choice, Walter's work is the result of testing how a characteristic brain resulted in conditioned behavior in a natural environment. To this end, he constructed three-wheeled automatons donned with lights that searched out other lights. They were analog devices, which used triodes and amplified feedback to mimic types of behavior. The feedback between the circuits caused the machine to display four distinct types of behavior qualified by the observer. The problem with Walter's analysis was that many of his assumptions were not tested outside of his own research. A profound discovery, however, is his notion of *free will* exhibited in his machines.

> "In his writings about the tortoises, Walter gave much weight to an attribute he called 'internal stability'[1] —the claimed ability of the tortoises to maintain their battery charge within limits by recharging themselves when necessary. A feature of the tortoises' circuitry was that, as the batteries became exhausted, the amplifier gain decreased, making it increasingly difficult to produce behavior pattern N (negative phototropism) [10].

Which is the attribute to avoid the light in the charging station where the feeding took place, this can be extended to match the main hypothesis presented in this document and pose the following experiment.

Method #3: If a machine has knowledge of its death, it fosters emergent behavior.

This theoretical assumption is simple: If a machine knows it can die, this knowledge and direct access in manipulating it physically will cause emergent behavior. Emergence is a property apart from the collection of quantification of its parts. The design of the artificial life system is a series of goal-based assumptions that guide the development of a methodology.

1. The first goal of a successful cybernetic system is that it should be fully autonomous. That is, once the system is started, it should require no further input from an external source or operator. In order to be autonomous, the system should be self-sufficient; i.e., have all the components necessary for its operation installed, variable in component and configuration depending on the expertise of the engineer. However, when the system is brought online, it should run continuously and without fail.
2. The second goal of a successful cybernetic system is that it should exist in situ or in context with its environment. It must be able to access experience from a native stimuli-response model with which to compose unique algorithms.
3. The third goal of a successful cybernetic system is that it should possess a system of behaviors relevant to its being. It should also have the ability to evolve and eventually reproduce.

---

[1] Walter borrowed this concept from his contemporary W. Ross Ashby who performed an exhaustive treatment of it he called the homeostat.

4. The fourth goal of a successful cybernetic system is that it must be in behavior indistinguishable from any other living system it mimics.

## The problem of choice

Can a machine make an aesthetic choice based on a non-determinate conditioning in an albeit restricted learning environment? There are a deterministic set of features, which can be obtained by quantifying behavior in the relativistic natural system considered for replication. Features present in the natural system can necessarily be considered for replication if they can be quantified into automata state-machine logic. Choice is narrowly defined as a pathway entity toward a goal-seeking behavior. Behavior is defined as a collection of choices requisite to a pathway solution dependent upon environmental factors weighted by a characteristically casual mechanism called free will. Testing these paradigms is irrelevant as these systems can be widely observed in nature; however, quantifying them into a software domain is a non-trivial task. There must be two features in tension against one another to allow choice:

1. A sensibility or "instinctual" aspect,
2. A means to create an understanding via transcendental analysis or "cognitive" aspect.

These two categories rely on the creation of events to flow from one state to another; these events are either triggered by external or internal stimuli. Each state is available for analysis at any linear time by reflection into the code blocks created as a function of the placement of them at the time of creation, and as a function of reaction to external stimuli. The culmination building a unique new feature, which replaces the previous state. Quantitatively, this is called the *experience*. Each experience is contrasted by the tension between the two categories based on range governance and through feedback, presents a cognitive workflow, which is a hybrid of the two original features. However, a caveat of the system is that the hybrid state cannot be subject to analysis within the host system; instead, it must be analyzed by impassive observation initialized by a third party resource.

This precludes that in order to study choice in quantified terms looking for the motivation of the choice and not the assemblage of choices themselves that several individualistic or autonomous systems must be harmoniously cooperating in some sort of orchestration. This eliminates most robotic systems now in use that look only to autonomy and social interaction; this also eliminates most of the adaptive machines currently under experimentation. Setting that hive-based automata state machines are the only structures able to withstand the conditions of stress in the environment and are subjected to the forces of evolution with as much import as their natural counterparts to human societies. A generalized model of robotic feeding and the application of choice is illustrated in Fig.2. Apart from strict considerations as a form of robust control or event-driven agents, robotic feeding considered here is analogous in form and function to an activity exhibited by organic entities, and can be reduced to a simple model of goal-seeking behavior.

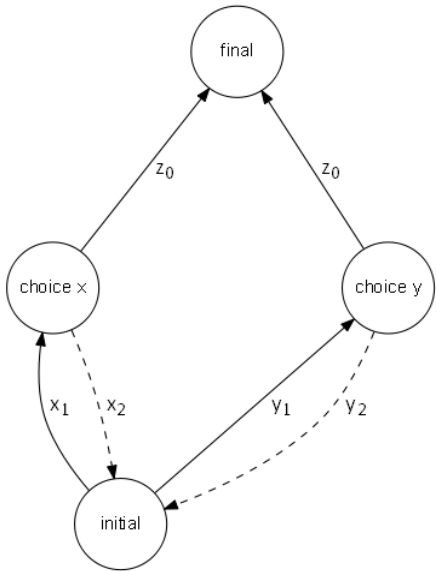

Fig.2. Activity of robotic feeding behavior.

A robot that is seeking power to recharge its onboard power system begins its activity at *initial* and is presented with one or more choices—in this example, two choices labeled $x$ and $y$—whereby to reach its necessary goal *final*. In order to decide which path to pursue, $x_1$ and $x_2$ toward $x$, $y_1$ and $y_2$ toward $y$, weights are assigned based on either success or failure of the path leading to the pursuit of *final* at $z_0$. Through repetition of this activity of seeking power, consecutive weights are averaged and the robot "prefers" pursuit of one path over the other because of positive experiences as well as negative feedback. Pseudocode of this activity is shown in Fig.3.

```
Activity of robotic feeling behavior - Pseudocode

    --Task--

    Recharge the batteries present in the system before power is exhausted.

    --Activity--

    Notice that the power level to sustain continuous operation is low enough to require recharge.
    Search stored data for available recharging types, of these types retrieve the weighted values
    to determine which is most optimal. If these values are equal to zero, generate a random number
    to choose which choice to pursue.

    Pursue choice 'x'. If recharging is reached, store a value of one for variable 'x1'; if
    recharging is not reached, store a value of zero for variable 'x1'.

    If recharging is not reached from pursuit of choice 'x', pursue choice 'y'. If recharging
    is reached, store a value of one for variable 'y1'; if recharging is not reached, store a
    value of zero for variable 'y1'.

    When the task is called, collect the weighted values for each recharging type. Sort the values
    in descending order. Pursue those choices on the list which have greater values. When pursing
    a choice by greater value and recharging is not reached, divide the value by two and store
    the new value.

    --End Activity--
```

Fig.3. Activity of robotic feeding behavior – Pseudocode.

The notion of the activity as a template for robotic feeding behavior serves as the primary theme for the description of the environment containing the robot. As such, the template can be expanded to include more detail relevant to ascribed behavior. In terms of defining a set of environmental factors, which facilitate the activity of power-seeking behavior, the process is modeled as a run-to-completion state machine, shown in Fig.4.

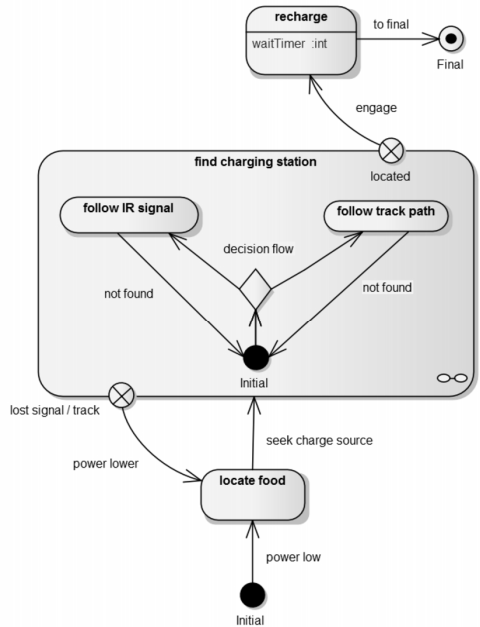

Fig.4. Finite-state diagram of robotic power-seeking behavior.

The components of Fig.4 depict the template in terms of a typical finite-state machine for a robot tasked with finding a recharging source. In ascribing behavior in an empathetic context, it is performing the task of searching for food. This activity is started when notified by the event *power low* wherein it will *locate food*. It will execute *seek charge source* entering the state machine.

The states, represented as boxes, are: *Initial*, *locate food*, *feeding*, and the state machine *find charging station* that contains *follow IR signal* and *follow track path*. The actions, represented as crossed circles, are two exits from the state machine. One is for a positive result, *located*, and one is for a negative result, *lost signal/track*. The transformations, represented as arrows: *power low*, *power lower*, *located*, are consequences of the choice following *seek charge source*. The transformation at the junction of *decision flow* indicates the decision since more than one outcome is present and the choice is made consequential of environmental factors. The software controlling the decision stipulates, without optimization, that it based on positive sensor feedback—if the IR signal or the track path is discovered first. The first acted upon, the alternative discarded unless the former returns a negative result.

The robot enters the state machine at the *Initial* orb when the sensor responsible for monitoring battery level notifies the operating system that power is low, noted in the transformation. When within the action *locate food*, a routine in the program executes the behavior for optimal seeking of a charge source. The transition of this behavior leads to entry into the *find charging station* state machine at *Initial*. If a positive result is obtained—that either of the choices are successful—the robot exits at *located*, and the transformation *engage* leads to the state *recharge*. When *waitTimer* expires, it will exit at *Final*. In terms of the complete behavior in this diagram, most of the complex behaviors are executed in the state machine, given the choice in the decision flow between to follow an infrared (IR) signal or follow a track path. Existence of such a choice is highly dependent on multiple solutions to the charging problem, if the state machine did not have both an IR source and track path to power to guide the robot, then choice in this context is irrelevant. In Fig.4, the experience derived from results of trying to follow one branch or another—found or not found—is one case of behavior. The experience derived from the pursuit of the specific choice—*located* or *lost signal/track*—is a second case. In the first case, not finding an IR source or a track path could be the result of neither existing nor unable to be found due to the causality of a sensor function designed to detect them. In the second case, having found the IR source or the track path but not locating it will keep motivation to continue finding it, or the robot remaining inside *find charging station*. When the state machine fails to return a positive result, it will exit at *lost signal/track*, the transformation then notes *power lower*, when compared to the transformation *power low*.

*Decision-making embedded within transformation logic*

Choice, in the scope detailed here, is a phenomenon isolated in the transformation between states yielding a consequence of one outcome. Given the power to select one outcome from many, the weight of consequence becomes determinate, e.g., one decision more optimal than another, to within a tolerance of 0.1 between weight values. From Fig.4, once the event for *locate food* is triggered. Fig.5 represents the behavior in the state machine *find charging station*.

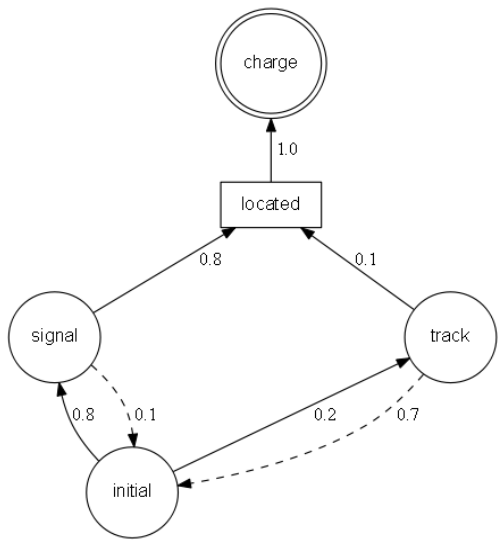

Fig.5. Weights for consequential decision-making.

The robot enters the diagram at *initial* and by reading the weights, can determine that recharging by using *signal* is better than *track* given the comparison positive weights (solid line) are 0.8 and 0.2, the comparison negative weights (dashed line) are 0.1 and 0.7, respectively. Within the context of the program, the weights for each decision path are averaged for each successful result. Each time the robot enters the *find charging station* state machine, it will learn to choose the optimal path because of the higher value of the weight. If decision paths have the same weight, a choice that is sufficiently random would assign a decision. The ethological implication of the modeled behavior embedded in the diagram of Fig.5 is the dynamic of it at different points of time during the activity of seeking a feeding source. The term "feeding" is applied here in the same scope as its original biological conception, that an entity pursuing food—in the case of the robot, energy—is exercising an adaptation for optimization of its survival.

According to Ashby [4, 11], states and their transformations are constructs of a characteristic map of behavior leading to the thinking process of entities, as noted here by the weights for each decision including positive and negative feedback. The implication is the fitness of the model and its completeness. What is illustrated in the runtime diagram are the degrees of change that the robot goes through during a finite quantity of time while attaining its goal. The model does not try to reveal the mechanisms behind the operations directly, rather, the character of the transition between states alluding to the behavior of the sequence. The goal is to reveal behavior of the robot during its power-seeking activity and gain evidence for the survival instinct in artificial systems.

A synopsis of what has been discussed thus far can be illustrated in Fig.6.

```
Synopsis:

A tasking presented to an entity possessing sensor items observing two objects and
free to exercise up to four momenta points in an environment.

Scenario:

ME I awake and find myself in darkness (white). 'Empty', I 'think' 'Zero, One, Two,
Three, and Four'. 'Search' to the left and the right, I 'find' move A and move B 'A
| B'. Search up and down, I find move C and move D 'C | D'. I find 'facilities'
sight, sound, touch. I 'see' a scene before me, a long plane extending 'forever': to
the left is a blue ball 'floating' and to the right is a red ball 'floating'. Is it
a dream?

Explicate:

'Empty' in that as a null item.
[white] as an arbitrary classification of an empty sensor network.
'think' as probing of the compiler architecture and routine structure.
'Zero, One, Two, Three, and Four' in that the degrees of freedom represented in the
programming framework [runtime].
'search' as probing hardware registry for a component.
'find' as a successful result of search.
'A | B' as encapsulated information received through a POST boot of output devices
connected.
'C | D' as encapsulated information received through a POST boot of output devices
connected.
'facilities' as a root descriptor for a sensor network comprised of singular members
working in unison [cooperation].
'see' in that based on feedback communication with sonic or optical sensors.
'forever' in that beyond the range of the sensor network.
'floating' in that it appears in a form that [logically] it should not.

[logically] in that it violates a law posited in the knowledge base.
```

Fig.6. Synopsis of a behavior designed to be a programmatic function.

## Scenario coding

Scenario coding is the process of a behavior trait into diagrams. The is to allow for the increasing levels of complexity realized in software programming. Scenarios, or algorithmic execution pathways, are similar to the creation of landscapes or circuit diagrams. Take the example of observing an object in the environment using a sensor or camera. The scenario begins with a textual description of what could happen if the sequence were to occur:

> ME I see two objects in my universe, one to the left and one to the right. ME I think about what they are and what I should do about them. Do I have a favorite? And do ME I have to make a decision?

Depicted graphically in Fig.7.

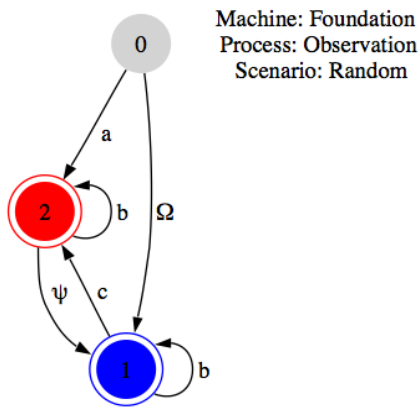

Fig.7. Observing environmental objects.

The scenario explained procedurally: The machine logic will begin at the grey circle marked '0' and note in program memory information about the two objects. The author is aware these behaviors are oversimplified and that it does not express the details of information exchange between sensors and the motor array not to mention language semantics in the event of the implementation of a communication exchange, to facilitate the analysis imbued by the behavior. Nevertheless, it demonstrates an empirical method to gather the necessary data *a priori* to classify which analytic to transcend to transform into information *a posteriori*. Another example, which demonstrates a choice made between a preference of two objects, begins with the scenario:

> ME I see two objects in my universe that I understand are variations on things that I like. ME I think about what I like about them and which is more important. What is my decision?

Where its representation is shown in Fig.8.

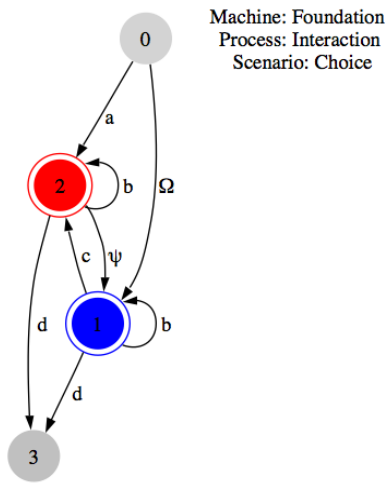

Fig.8. Arriving at a choice between objects.

The scenario explained procedurally: The machine logic will begin at the grey circle marked '0' and express an output at the grey circle marked '3'. This scenario indicates that the inner workings of the algorithm exhibited in the drawing will work to generate an external manifestation whereas in the previous example, the entire procedure is compartmental.

# 4      Conclusions

This document discussed the method of artificial systems. While not claiming any authority on the matter, it was the hope of the author to persuade the reader by intellectual argument and philosophical traceability that it would finally be possible to classify and expect the kinds of behaviors in systems imbued with artificial intelligence. An anticipated outcome of this work would be to advise governmental bodies as to the type of ethical construct which can be fairly applied to keep creativity in robotic programming, while introducing a set of safeguards to help keep the program from going rogue.

* * * * * * * *